\documentclass[10pt,letterpaper]{article}
\usepackage[top=0.85in,left=2.75in,footskip=0.75in]{geometry}

\usepackage{amsmath,amssymb}

\usepackage{changepage}

\usepackage{booktabs}

\usepackage{textcomp,marvosym}

\usepackage{cite}

\usepackage{nameref,hyperref}

\usepackage[right]{lineno}

\usepackage[nopatch=eqnum]{microtype}
\DisableLigatures[f]{encoding = *, family = * }

\usepackage[table]{xcolor}

\usepackage{array}

\newcolumntype{+}{!{\vrule width 2pt}}

\newlength\savedwidth

\raggedright
\usepackage[aboveskip=1pt,labelfont=bf,labelsep=period,justification=raggedright,singlelinecheck=off]{caption}

\makeatletter
\renewcommand{\@biblabel}[1]{\quad#1.}
\makeatother

\usepackage{lastpage,fancyhdr,graphicx}
\usepackage{epstopdf}
\fancyheadoffset[L]{2.25in}
\fancyfootoffset[L]{2.25in}
\begin{document}
\vspace*{0.2in}

\begin{flushleft}
{\Large
\textbf\newline{FedStrategist: A Meta-Learning Framework for Adaptive and Robust Aggregation in Federated Learning} 
}
\newline
\\
Md Rafid Haque\textsuperscript{1*},
Dr. Abu Raihan Mostofa Kamal\textsuperscript{1},
Dr. Md. Azam Hossain\textsuperscript{1}
\\
\bigskip
\textbf{1} Department of Computer Science and Engineering, Islamic University of Technology (IUT), Boardbazar, Gazipur - 1704, Bangladesh.

\bigskip

* rafidhaque@iut-dhaka.edu
\bigskip







\end{flushleft}
\section*{Abstract}
Federated Learning (FL) offers a paradigm for privacy-preserving collaborative AI, but its decentralized nature creates significant vulnerabilities to model poisoning attacks. While numerous static defenses exist, their effectiveness is highly context-dependent, often failing against adaptive adversaries or in heterogeneous data environments. This paper introduces FedStrategist, a novel meta-learning framework that reframes robust aggregation as a real-time, cost-aware control problem. We design a lightweight contextual bandit agent that dynamically selects the optimal aggregation rule from an arsenal of defenses based on real-time diagnostic metrics. Through comprehensive experiments, we demonstrate that no single static rule is universally optimal. We show that our adaptive agent successfully learns superior policies across diverse scenarios, including a "Krum-favorable" environment and against a sophisticated "stealth" adversary designed to neutralize specific diagnostic signals. Critically, we analyze the paradoxical scenario where a non-robust baseline achieves high but compromised accuracy, and demonstrate that our agent learns a conservative policy to prioritize model integrity. Furthermore, we prove the agent's policy is controllable via a single "risk tolerance" parameter, allowing practitioners to explicitly manage the trade-off between performance and security. Our work provides a new, practical, and analyzable approach to creating resilient and intelligent decentralized AI systems.


\section{Introduction}
Federated Learning (FL) has emerged as a paradigm-shifting approach to machine learning, enabling collaborative model training on decentralized data sources without compromising user privacy \cite{mcmahan2016communication, li2019federated}. By keeping raw data localized on user devices or within organizational silos, FL inherently addresses critical privacy concerns and regulatory hurdles like GDPR, making it a key enabling technology for sensitive domains such as healthcare and finance \cite{rahman2021challenges, zeydan2023self}. However, this very decentralization creates a trust vacuum. In the absence of a central, trusted authority with visibility into local data, FL systems are vulnerable to a spectrum of adversarial behaviors.

The security challenges are significant and well-documented. Malicious participants can launch poisoning attacks, manipulating their model updates to degrade the global model's performance \cite{fang2019local, shejwalkar2021manipulating} or insert targeted backdoors \cite{bagdasarian2018how}. Furthermore, the collaborative model creates incentive problems, leading to free-riding, where clients benefit from the global model without contributing useful work \cite{lin2019freeriders}.

To address these vulnerabilities, researchers have turned to blockchain technology as a foundation for building decentralized trust \cite{li2021blockchain, ghanem2021flobc}. The blockchain's promise of an immutable, transparent, and auditable ledger provides a robust substrate for coordinating FL processes and enforcing protocol rules. This has led to the emergence of Blockchain-based Federated Learning (BCFL), a vibrant field aiming to create fully decentralized and trustworthy AI ecosystems \cite{goh2023blockchain, shayan2021biscotti}.

The core challenge in BCFL is to design mechanisms that ensure participants act honestly. The literature reveals a diverse ecosystem of solutions built upon three pillars of trust: \textbf{reputation}, \textbf{incentives}, and \textbf{verification}. Reputation systems track past behavior to build social trust \cite{aluko2021proof, nasrulin2022meritrank}; incentive and tokenomic models use economic principles to align self-interest with collective goals \cite{lamberty2020efficiency, zhan2021survey}; and cryptographic verification systems use mathematics to provide provable guarantees of honesty \cite{xing2023zero, zhang2025zk}.

However, despite this wealth of solutions, most defenses are implemented as \textit{static, context-insensitive mechanisms}. A fundamental finding, which we confirm in this paper, is that no single static aggregation rule is universally optimal. A geometrically-motivated defense like Krum, for example, is effective against simple outlier attacks but fails completely when its assumptions are violated by high data heterogeneity \cite{blanchard2017machine}. This brittleness of static defenses necessitates a move towards dynamic, adaptive strategies. While some adaptive frameworks exist, they often introduce significant computational complexity or rely on deterministic policies that can be exploited by sophisticated, second-order adversaries \cite{krauss2024automatic, qian2024developing}.

This paper addresses this critical gap. We introduce \textbf{FedStrategist}, a novel meta-learning framework that reframes robust aggregation as a real-time, cost-aware control problem. We propose a lightweight contextual bandit agent that learns a dynamic policy for selecting the optimal aggregation rule from an arsenal of defenses. The agent's decision is guided by a real-time analysis of the network's heterogeneity and threat state, and its policy is trained to explicitly manage the trade-off between model accuracy and the computational cost of security. Through extensive experiments, we demonstrate that our agent learns to outperform any single static defense across diverse environments, successfully adapts its strategy to different attack vectors, and remains robust even when its diagnostic signals are being actively manipulated. Our work provides a practical, tunable, and analyzable framework for creating the next generation of resilient and intelligent decentralized AI systems.

\section{Foundational Concepts}
\label{sec:foundations}
To construct our taxonomy, we must first establish a common understanding of the core technologies and the challenges that motivate their integration. 

\subsection{Federated Learning and its Inherent Vulnerabilities}
Federated Learning (FL) is a distributed machine learning paradigm designed to train a model on decentralized data \cite{mcmahan2016communication, yurdem2024federated, hu2021federated}. In the canonical FedAvg algorithm, a server orchestrates training rounds where clients compute updates on local data, which are then averaged by the server \cite{konecny2016federated, rtinsights2024federated}. The primary benefit is privacy, as raw data never leaves the client's device. However, this creates several challenges, including high communication costs \cite{sensors2023communication, le2024exploring, lan2023communication}, and systemic issues arising from data and systems heterogeneity \cite{zhao2018federated, gao2022survey, fang2023heterogeneous}.

Most critically, the lack of server visibility creates a fundamental trust deficit, making FL vulnerable to a spectrum of adversarial attacks \cite{shejwalkar2021manipulating}. These attacks can be broadly categorized as:
\begin{itemize}
    \item \textbf{Poisoning Attacks:} Malicious clients can intentionally submit corrupted model updates. Untargeted (Byzantine) attacks aim to degrade overall performance \cite{blanchard2017machine}, while targeted (backdoor) attacks aim to make the model misclassify specific inputs chosen by the attacker \cite{bagdasarian2018how}.
    \item \textbf{Free-Riding and Collusion:} Malicious or lazy clients can benefit from the final trained model without contributing useful work \cite{lin2019freeriders}. Moreover, adversaries can coordinate their attacks in a collusive manner, amplifying their impact and evading defenses designed to spot individual outliers \cite{fung2018foolsgold, feng2025dmpa, ranjan2022securing}.
    \item \textbf{Privacy Leakage:} Despite keeping data local, the shared gradients themselves can be exploited to reconstruct sensitive private data, a vulnerability famously demonstrated by Deep Leakage from Gradients \cite{zhu2019deep}.
\end{itemize}
These inherent vulnerabilities highlight the need for a more robust framework that can establish trust, verify behavior, and incentivize honesty in a decentralized setting.

\subsection{Blockchain and Smart Contracts as a Trust Foundation}
Blockchain technology offers a compelling solution to the trust deficit in FL. At its core, a blockchain is an immutable, distributed ledger maintained by a network of nodes without a central authority \cite{li2021blockchain, ghanem2021flobc}. Agreement on the state of the ledger is achieved through a \textbf{consensus protocol}, a topic extensively surveyed in \cite{xiao2020survey, alkhodair2023consensus}. While early protocols like Proof-of-Work (PoW) are computationally expensive, modern systems often employ variants of \textbf{Byzantine Fault Tolerance (BFT)}.

BFT protocols, originating from the theoretical Byzantine Generals Problem \cite{lamport1982byzantine}, are designed to allow a distributed system to reach agreement even if a fraction of participants are malicious. Practical Byzantine Fault Tolerance (PBFT) \cite{castro1999practical} and its modern, more scalable successors like HotStuff \cite{yin2019hotstuff} provide the deterministic finality and operational resilience needed for complex coordination tasks like FL.

This foundation is brought to life by \textbf{smart contracts}: self-executing programs whose code is stored on the blockchain. These contracts can automate the rules of an FL protocol, but their security is paramount, as vulnerabilities can lead to catastrophic losses \cite{praitheeshan2019security}.

\subsection{The Pillars of Trust in Decentralized AI}
Building upon the blockchain foundation, the research community has developed three primary mechanisms, or "pillars", to construct trustworthy decentralized AI systems:
\begin{enumerate}
    \item \textbf{Reputation and Identity Systems:} These mechanisms quantify trustworthiness based on historical behavior \cite{aluko2021proof, nasrulin2022meritrank}. This includes general reputation models \cite{ren2025beyond} and the use of \textbf{Decentralized Identity (DID)} to provide a persistent, self-sovereign anchor for reputation, mitigating Sybil attacks \cite{geng2021did, zeydan2023self}.
    \item \textbf{Incentive Mechanisms \& Tokenomics:} Addressing the "why" of participation, these systems use economic principles from game theory and mechanism design to create token-based economies \cite{lamberty2020efficiency, ito2024cryptoeconomics, aamas2012token, tdefi2024gametheory, nextrope2024applying, linkedintokenomics2024}. By rewarding desirable actions and punishing malicious behavior, they aim to make honesty the most profitable strategy \cite{zhan2021survey, coinsquare2025gametheory, murano2025incentive, encyclopedia2024gametheory}.
    \item \textbf{Cryptographic Verification:} This pillar replaces long-term trust with immediate, mathematical proof of correctness. It encompasses \textbf{Zero-Knowledge Proofs (ZKPs)}, which prove honest computation without revealing private data \cite{zhang2025zk, xing2023zero, acar2017survey}, and \textbf{Homomorphic Encryption (HE)}, which allows computation on encrypted data \cite{han2025pbfl, yang2023review}. These tools, while powerful, often come at a significant computational cost \cite{heiss2022advancing, kokaj2025mathematical}.
\end{enumerate}

\begin{figure}[htbp]
    \centering
    \includegraphics[width=0.9\columnwidth]{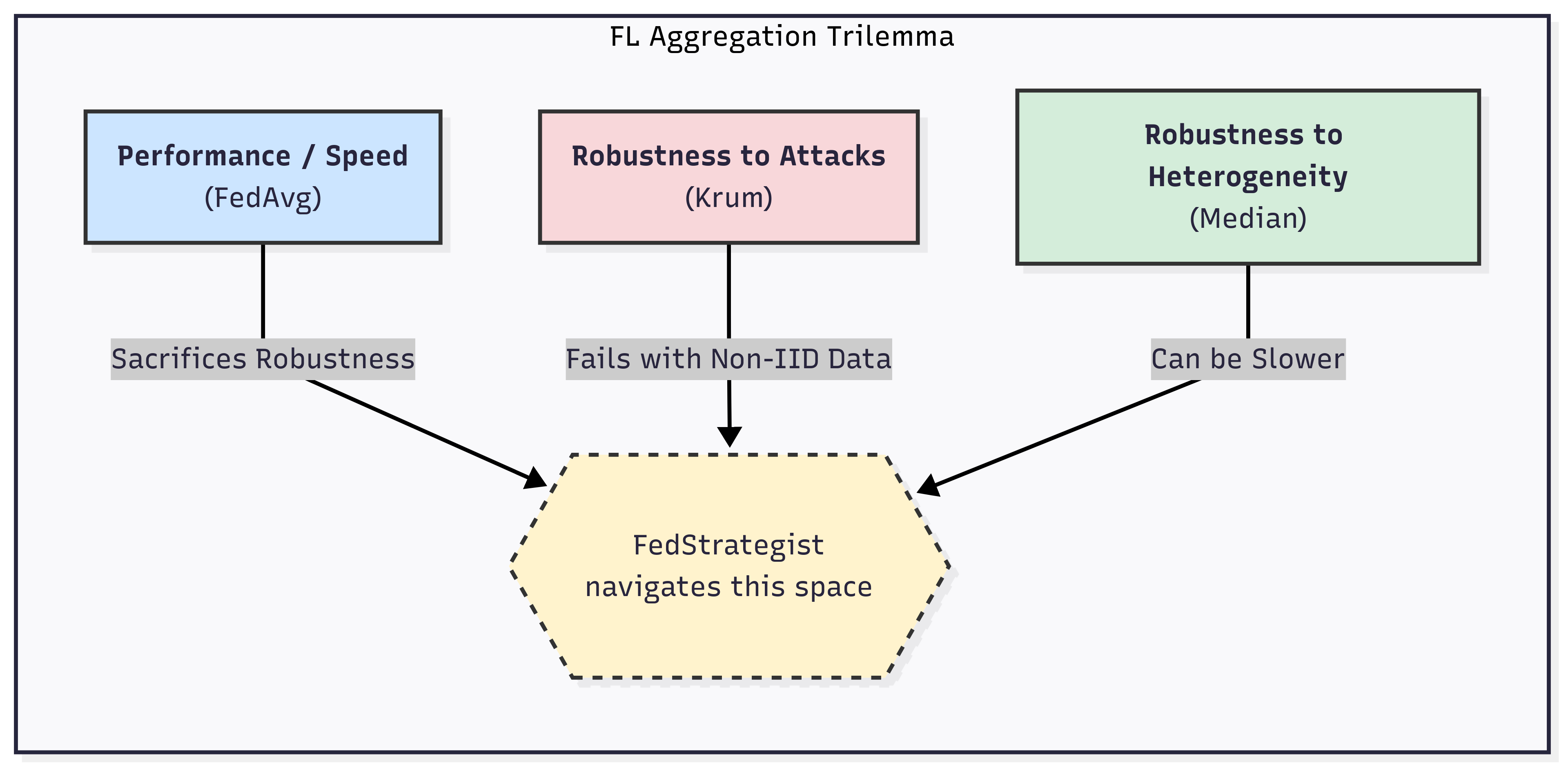}
    \caption{The Federated Learning Aggregation Trilemma. Static aggregation rules force a trade-off between performance (FedAvg), robustness to attacks (Krum), and robustness to data heterogeneity (Median). The FedStrategist framework is designed to dynamically navigate this trade-off space by selecting the optimal rule for the current network conditions.}
    \label{fig:trilemma}
\end{figure}

These distinct approaches to trust highlight a fundamental challenge in designing robust FL systems, which we conceptualize as the "Aggregation Trilemma," illustrated in Fig.~\ref{fig:trilemma}. A practitioner must often choose between optimizing for performance, robustness to attacks, or robustness to data heterogeneity, as no single static rule excels at all three. This motivates the need for an adaptive framework capable of navigating this trade-off space in real time.

\section{Related Work}
\label{sec:related_work}

Our research is situated at the confluence of three major research streams: 1) robust aggregation mechanisms in federated learning, 2) adaptive and dynamic defense strategies, and 3) the application of reinforcement learning for resource management in distributed systems. This section surveys these domains to precisely situate our contribution as a necessary advancement over the current state-of-the-art.

\subsection{Static Defenses and Their Inherent Limitations}
The foundational challenge in securing Federated Learning (FL) is defending the global model from malicious updates submitted by Byzantine clients \cite{fang2019local, shejwalkar2021manipulating}. A significant body of work has focused on developing "robust" aggregation rules to replace the vulnerable Federated Averaging (FedAvg) \cite{mcmahan2016communication} algorithm. These static defenses can be broadly grouped into two families.

The first relies on robust statistics to identify a central tendency resistant to outliers, such as Coordinate-wise Median and Trimmed Mean \cite{yin2018byzantine, luan2024robust}. The second uses geometric heuristics, most notably Krum, which selects updates based on their proximity to neighboring updates in the parameter space \cite{blanchard2017machine}.

However, a critical and well-documented weakness of all static defenses is their predictability and context-insensitivity. Research has shown that their fixed logic can be circumvented by sophisticated adversaries \cite{fang2019local}. Furthermore, their effectiveness is highly dependent on the underlying data distribution; for instance, Krum's performance degrades significantly in non-IID settings where the updates from honest clients are naturally far apart \cite{li2024experimental}. This inflexibility motivates the need for dynamic and adaptive defense strategies.

\subsection{Adaptive Defenses: From Rule-Switching to Contextual Awareness}
Recognizing the limitations of static defenses, a second wave of research has focused on creating adaptive systems. The pioneering work in this area is the SARA framework by Hu et al. \cite{sara2025adapting}, which was the first to formalize aggregation rule adaptation as a Multi-Armed Bandit (MAB) problem. In SARA, each aggregation rule is an "arm," and a UCB algorithm selects the optimal rule based on its historical reward, demonstrating the viability of this approach against dynamic attack patterns.

While foundational, this context-free MAB approach has significant limitations. A standard UCB agent learns the best overall strategy but cannot react to the specific conditions of a single round. It is like a doctor prescribing an antibiotic based on its average success rate over the past year, without considering the patient's immediate symptoms. This approach can be slow to adapt and is vulnerable to adversaries who can manipulate the environment in ways that make the long-term average reward a poor predictor of short-term success.

Our work, FedStrategist, directly addresses this gap. We advance the state-of-the-art by moving from a simple MAB to a contextual bandit. The core novelty of our framework is the introduction of a lightweight, real-time diagnostic state vector, $S_t$. This allows our agent to learn a much richer policy, $\pi(S_t) \to A_j$, that maps the current symptoms of the network to the optimal defensive action. This shift from a context-free to a context-aware agent is a critical step towards creating a more intelligent and responsive defense.

\subsection{Reinforcement Learning for Optimization in FL}
The third relevant domain is the application of reinforcement learning (RL) and bandits to optimize processes in FL. This approach has proven effective for resource management challenges like client selection, where systems like MAB-RFL and UCB-CS use bandits to select the most promising clients for participation \cite{wan2021mabrfl, cho2023ucb}.

Most relevant to our thesis is the "FedCostAware" framework by Sinha et al., which introduces a cost-benefit reward function to manage the financial expense of FL on cloud infrastructure \cite{sinha2025fedcostawareenablingcostawarefederated}. Their work establishes the principle of using a tunable parameter, $\lambda$, to explicitly manage trade-offs.

Our FedStrategist framework creates a novel synthesis of these research streams. While SARA introduced the MAB for rule selection, its approach is context-free and does not consider the cost of defense. While FedCostAware introduced cost-awareness, it was for client scheduling, not defensive strategy. We are the first to unify these concepts. We propose a lightweight contextual bandit that explicitly and dynamically manages the cost-versus-robustness trade-off for the specific task of aggregation rule selection, and provide a deep analysis of the resulting learned policy and its stability. It is at this precise intersection—a practical, tunable, and context-aware controller for federated aggregation—that our work makes its novel contribution.

\section{Materials and Methods}
\label{sec:materials_methods}

\begin{figure}[ht]
    \centering
    \includegraphics[width=\textwidth]{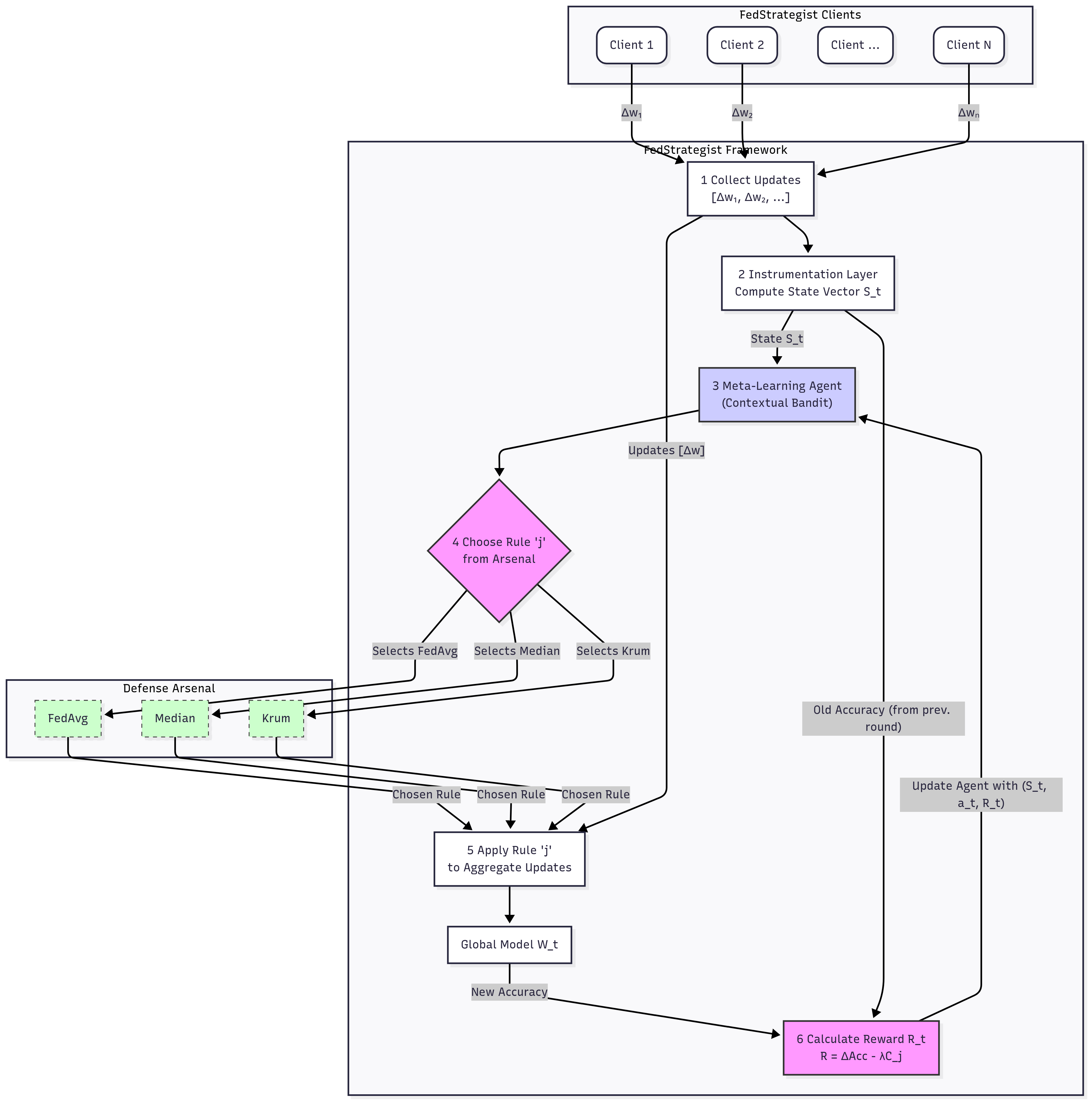}
    \caption{The FedStrategist Framework Architecture. In each round ($t$), (1) the server collects updates ($\Delta w_i$) from clients. (2) The Instrumentation Layer computes a state vector ($S_t$). (3) The Meta-Learning Agent (Contextual Bandit) ingests $S_t$ and (4) chooses an aggregation rule '$j$' from the Defense Arsenal. (5) The chosen rule is applied to the updates to compute the new global model $W_t$. (6) A reward $R_t$ is calculated based on the resulting accuracy change and the rule's cost, which is then used to update the agent's policy.}
    \label{fig:architecture}
\end{figure}

Our research introduces \textit{FedStrategist}, a meta-learning framework designed to dynamically select the optimal aggregation rule in Federated Learning (FL). As depicted in Fig.~\ref{fig:architecture}, the system operates as a \textbf{closed-loop control system}. In each round, client updates are first analyzed by an instrumentation layer to produce a diagnostic state vector. This state informs the real-time selection of an aggregation rule by a contextual bandit agent. The resulting model performance is then used to generate a reward signal, which updates the agent's policy for future rounds. This section details the theoretical underpinnings of each component of this architecture.

\subsection{Federated Learning Framework}
Our simulation is built upon the canonical Federated Averaging (FedAvg) algorithm \cite{mcmahan2016communication}. The system consists of a central server and a set of $N$ clients. In each communication round $t$, the following process occurs:
\begin{enumerate}
    \item The server distributes the current global model state, $W_t$, to all participating clients.
    \item Each client $i$ updates the model using its local private dataset, $D_i$, for $E$ local epochs, producing a local model update, $\Delta w_i^t$.
    \item The clients transmit their updates back to the server.
    \item The server applies a dynamically selected aggregation rule, $A_j$, to the set of client updates $\{\Delta w_i^t\}$ to compute the new global model, $W_{t+1}$.
\end{enumerate}

\subsection{The Defense Arsenal: Aggregation Rules}
The core of our framework is a selectable arsenal of aggregation rules. \textbf{We selected three archetypal algorithms to represent the major families of defense strategies: non-robust, statistical, and geometric.}
\begin{itemize}
    \item \textbf{Federated Averaging (FedAvg):} The standard, non-robust baseline, which computes the element-wise mean:
    \[ W_{t+1} = \frac{1}{N} \sum_{i=1}^{N} \Delta w_i^t \]
    \item \textbf{Coordinate-wise Median:} A statistically robust method that computes the median for each individual parameter across all client updates, providing strong resilience to value-based outliers.
    \item \textbf{Krum:} A geometrically robust method that defends against a fraction $f$ of Byzantine clients \cite{blanchard2017machine}. It selects the single client update that minimizes the sum of squared Euclidean distances to its $k = N - f - 2$ nearest neighbors, assuming honest clients form a tight geometric cluster.
\end{itemize}

\subsection{Adversarial Threat Model}
To evaluate the robustness of our framework against different threat profiles, we implement two distinct model poisoning attacks, inspired by the work of Fang et al. \cite{fang2019local}:
\begin{enumerate}
    \item \textbf{Standard Poisoning Attack:} A "loud" attack where the malicious client scales its benign update direction by a large factor (e.g., $5\times$), creating a high-magnitude update designed to be easily detectable by norm-based defenses but highly disruptive to mean-based aggregators.
    \item \textbf{Stealth Poisoning Attack:} An "intelligent" attack where the adversary crafts its malicious update to precisely match the average L2 norm of the benign clients from the current round, making it invisible to defenses that rely on detecting magnitude-based outliers.
\end{enumerate}

\subsection{Instrumentation: The Diagnostic State Vector}
To enable adaptive decision-making, the server computes a low-dimensional state vector, $S_t$, from the set of client updates. \textbf{These metrics were chosen to be computationally lightweight while capturing diverse information about the geometry and distribution of the updates.} Our state vector $S_t \in \mathbb{R}^3$ is composed of:
\begin{enumerate}
    \item \textbf{Variance of Update Norms:} The statistical variance of the L2 norms of all client update vectors. A high variance suggests the presence of magnitude-based outliers.
    \item \textbf{Average Pairwise Cosine Similarity:} The mean cosine similarity computed between all pairs of client update vectors. This captures the degree of directional agreement among clients.
    \item \textbf{Mean Update Norm:} The L2 norm of the globally averaged update vector, $\|\frac{1}{N}\sum_i \Delta w_i^t\|_2$, which captures the magnitude of the collective step.
\end{enumerate}

\subsection{The Commander: A Contextual Bandit for Rule Selection}
The core of FedStrategist is a meta-learning agent that learns a policy, $\pi: S_t \to A_j$, to map the current state vector to an optimal aggregation rule. We model this as a contextual bandit problem and implement the agent using the LinUCB algorithm.

For each action (aggregation rule) $a$, LinUCB models the expected reward as a linear function of the context $x_t$ (our state vector $S_t$): $E[r_t|x_t, a] = x_t^T \theta_a$. In each round, it chooses the action that maximizes the Upper Confidence Bound (UCB) on this estimate:
\[ \text{action}_t = \arg\max_{a \in \mathcal{A}} (x_t^T \hat{\theta}_a + \alpha \sqrt{x_t^T A_a^{-1} x_t}) \]
where $\hat{\theta}_a$ is the estimated coefficient vector for action $a$, $A_a$ is a matrix updated from the training data, and $\alpha$ is an exploration parameter.

The agent is trained using a reward signal, $R_t$, that explicitly models the trade-off between model improvement and the computational cost of the defense:
\[ R_t = (\text{Acc}_{t} - \text{Acc}_{t-1}) - (\lambda_{cost} \times C_j) \]
where the accuracy is measured on a small, server-held proxy validation set, $C_j$ is a normalized heuristic cost for the chosen rule $j$, and $\lambda_{cost}$ is the key hyperparameter that defines the system's "risk tolerance."

\section{Environment Setup and Implementation}
\label{sec:implementation}

All experiments were conducted using a custom simulation framework built in Python 3.9 with the PyTorch 1.12 library. Computations were performed on a single NVIDIA RTX 4050 GPU with 6GB of VRAM.

\subsection{Dataset and Data Partitioning}
We use the CIFAR-10 dataset for all experiments. To simulate statistical data heterogeneity, we partitioned the training set among $N$ clients using a Dirichlet distribution, controlled by the concentration parameter $\beta$. A small $\beta$ (e.g., 0.1) results in a highly non-IID distribution where clients may only have data from a few classes. A large $\beta$ (e.g., 10.0) results in a nearly IID distribution. This method allows us to precisely control the level of data heterogeneity, a critical variable in our experiments.

\subsection{Model Architecture and Training}
For the FL task, we use a simple Convolutional Neural Network (CNN) architecture, detailed in our open-source repository. Each client trains the model on its local data for $E=1$ epoch using Stochastic Gradient Descent (SGD) with a learning rate of 0.001 and momentum of 0.9.

\subsection{Framework Implementation}
Our simulation framework consists of several key modules:
\begin{itemize}
    \item \textbf{'fl\_core.py'}: Contains the 'Server' and 'Client' classes. The 'Server' class manages the global model, orchestrates rounds, and houses the bandit agent. The 'Client' class handles local training. A 'MaliciousClient' subclass inherits from 'Client' and overrides the training method to inject attacks.
    \item \textbf{'aggregation.py'}: Implements the "Defense Arsenal" as a set of standalone functions ('fed\_avg', 'coordinate\_wise\_median', 'krum').
    \item \textbf{'attacks.py'}: Contains the logic for the "standard" and "stealth" poisoning attacks.
    \item \textbf{'diagnostics.py'}: Implements the functions to compute the state vector metrics.
    \item \textbf{'bandit.py'}: Contains a standard implementation of the LinUCB algorithm using NumPy and Scikit-learn's Ridge regression for stable linear modeling.
\end{itemize}
The entire codebase and experimental configurations are permanently archived in \cite{haque_2025} to ensure full reproducibility of our results.

\section{Results}
\label{sec:results}

To evaluate our proposed framework, FedStrategist, we conducted a series of experiments designed to systematically test its performance and adaptability under diverse conditions of data heterogeneity and adversarial threat. This section presents the empirical results that validate our core hypotheses.

\subsection{Conceptualizing the Learned Policy}
\begin{figure}[htbp]
    \centering
    \includegraphics[width=0.9\columnwidth]{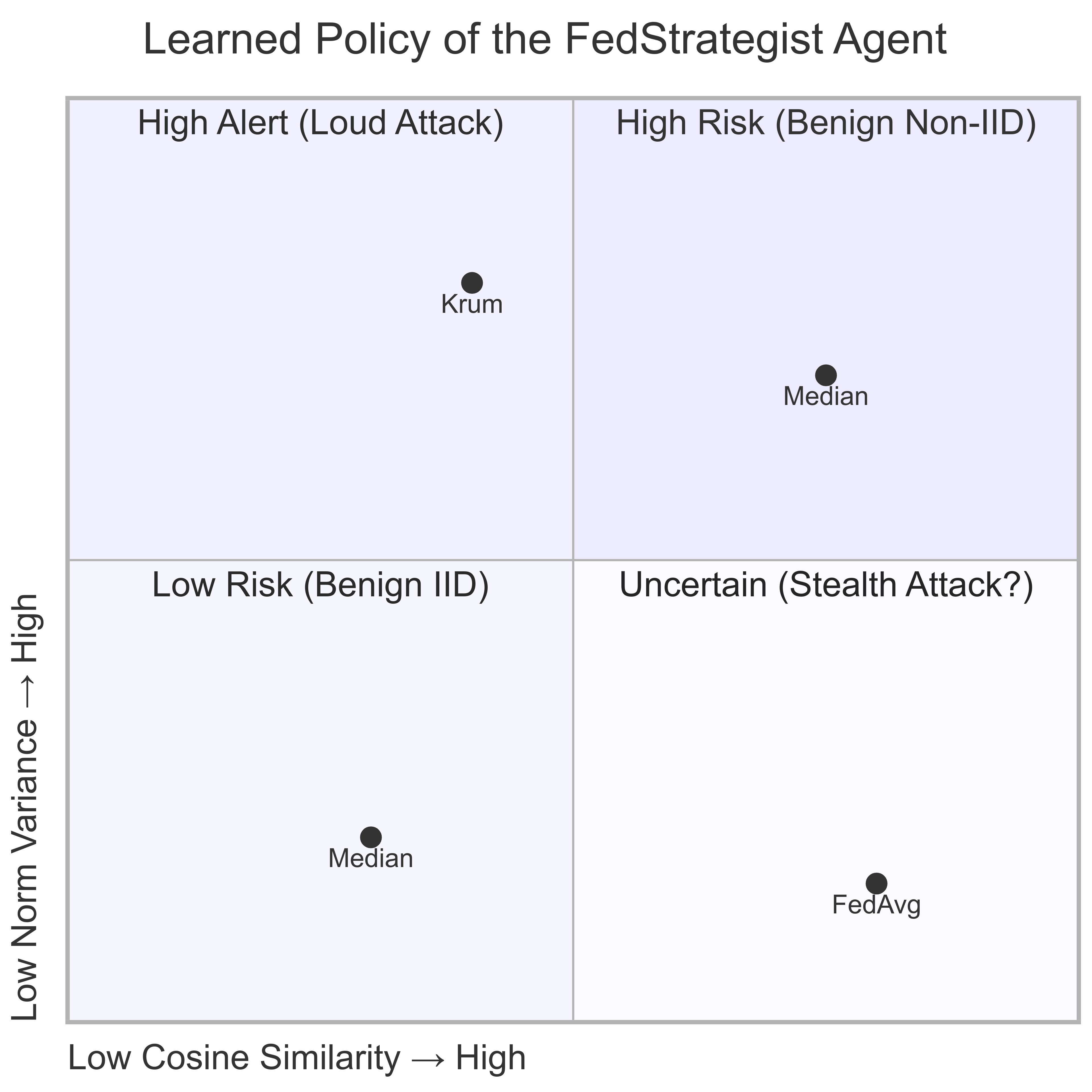}
    \caption{Conceptual Visualization of the Learned Policy. This diagram illustrates the idealized decision regions. The agent is expected to learn to select the aggressive 'FedAvg' in safe conditions (low variance, high similarity), the geometric 'Krum' against loud attacks (high variance), and the statistical 'Median' when faced with high data heterogeneity (low similarity).}
    \label{fig:policy_conceptual}
\end{figure}

Before presenting the quantitative results, it is instructive to visualize the policy our agent is expected to learn. The agent's task is to map the diagnostic state space, primarily defined by update similarity and variance, to an optimal action. Figure~\ref{fig:policy_conceptual} illustrates the idealized decision regions we hypothesize a successful agent will learn, forming the basis for our experimental validation.

\subsection{Performance under a Standard Poisoning Attack}
Our initial experiment established a performance baseline against a standard, high-magnitude model poisoning attack. We evaluated four strategies—our 'adaptive' agent versus static 'FedAvg', 'Median', and 'Krum'—across high ($\beta=0.1$) and moderate ($\beta=0.5$) data heterogeneity, with 5 malicious clients out of 20.

\begin{figure}[htbp]
    \centering
    \includegraphics[width=\columnwidth]{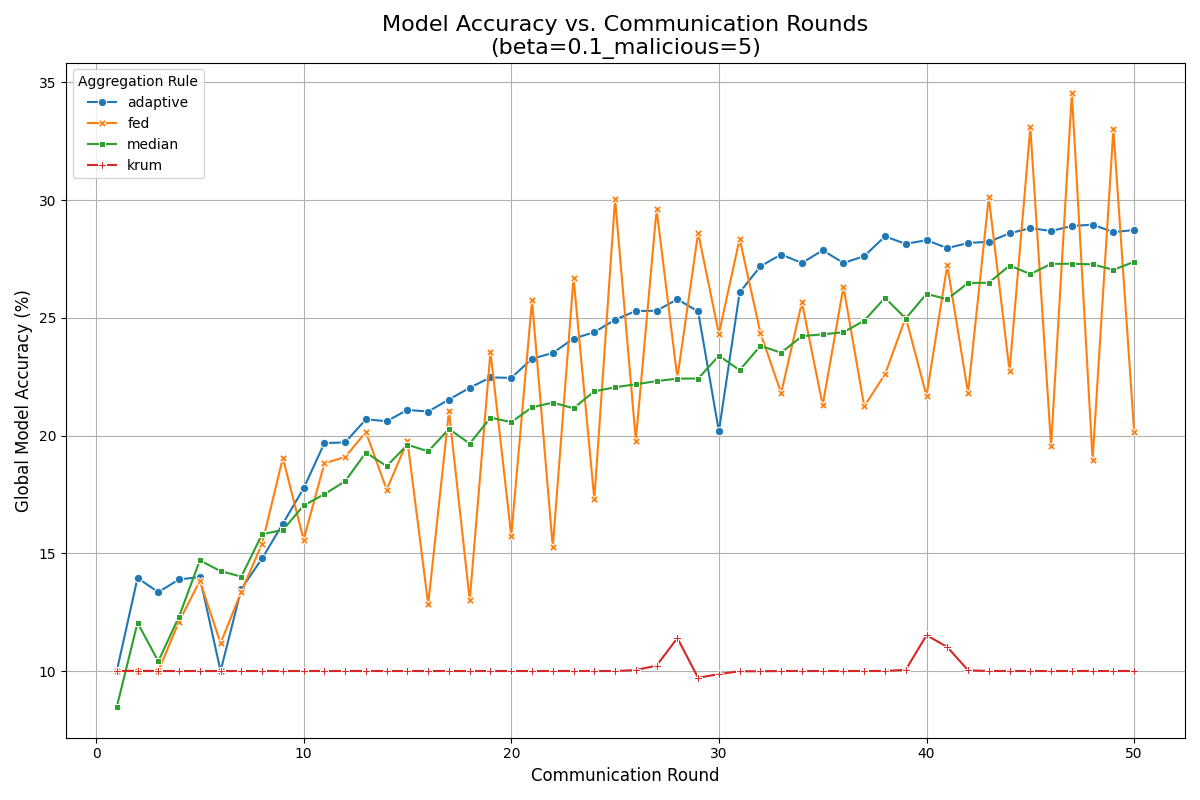} 
    \includegraphics[width=\columnwidth]{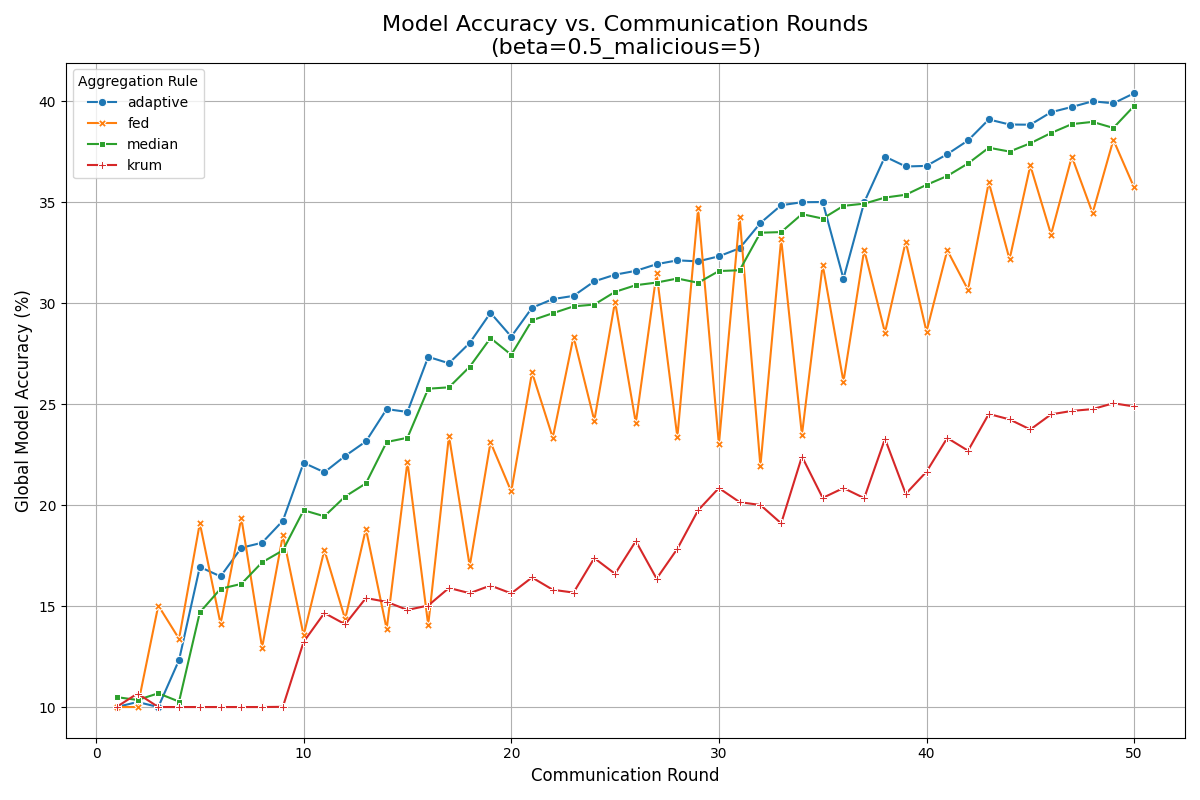} 
    \caption{Model accuracy under a standard poisoning attack. Top: High heterogeneity ($\beta=0.5$). Bottom: Moderate heterogeneity ($\beta=0.5$). The FedStrategist ('adaptive') agent learns a policy that surpasses all static baselines.}
    \label{fig:exp_standard_attack}
\end{figure}

The results, presented in Fig.~\ref{fig:exp_standard_attack}, yield two key findings. First, no single static rule is universally optimal. 'Krum' fails completely in the high-heterogeneity environment, as its geometric assumptions are violated. Second, our 'adaptive' agent successfully navigates this trade-off. In both scenarios, it learns to avoid the unstable 'FedAvg' and the ineffective 'Krum', identifying 'Median' as the most reliable static baseline and ultimately achieving superior performance by learning an optimal switching policy.

\subsection{Robustness to an Adaptive Stealth Adversary}
The second experiment tested the system's resilience against a more sophisticated adversary. We introduced a "stealth" poisoning attack where malicious clients disguise their updates by normalizing their magnitude, specifically targeting norm-based diagnostic metrics.

\begin{figure}[htbp]
    \centering
    \includegraphics[width=\columnwidth]{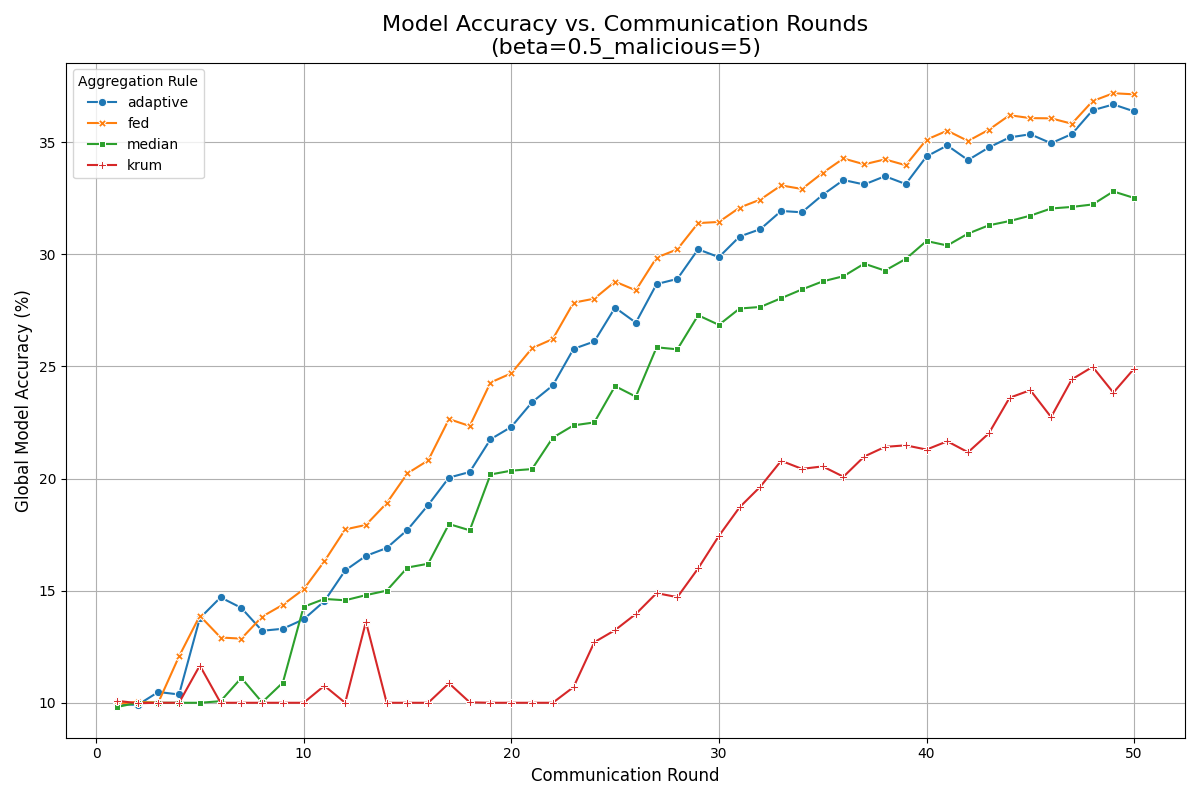}
    \caption{Model accuracy under a stealth poisoning attack ($\beta=0.5$). The attack's regularizing effect creates a complex trade-off between the compromised 'FedAvg' and the safer robust methods.}
    \label{fig:exp_stealth_attack}
\end{figure}

The results, shown in Fig.~\ref{fig:exp_stealth_attack}, are highly insightful. The attack's regularizing effect caused the non-robust 'FedAvg' to achieve the highest raw accuracy. However, this model is fundamentally compromised. The key result is that our 'adaptive' agent, despite one of its primary diagnostic metrics being "blinded," still learned a policy that outperformed the best static robust aggregator ('Median'). It successfully learned to rely on its remaining signals (like cosine similarity) to make intelligent decisions, demonstrating the resilience of our multi-metric state representation.

\subsection{Validation of Generalist Strategy in a Krum-Favorable Environment}
Our third experiment was designed to test the agent's generality and lack of inherent bias. We constructed a scenario theoretically optimal for \texttt{Krum}: low data heterogeneity ($\beta=10.0$) and a "loud" standard attack.

\begin{figure}[htbp]
    \centering
    \includegraphics[width=\columnwidth]{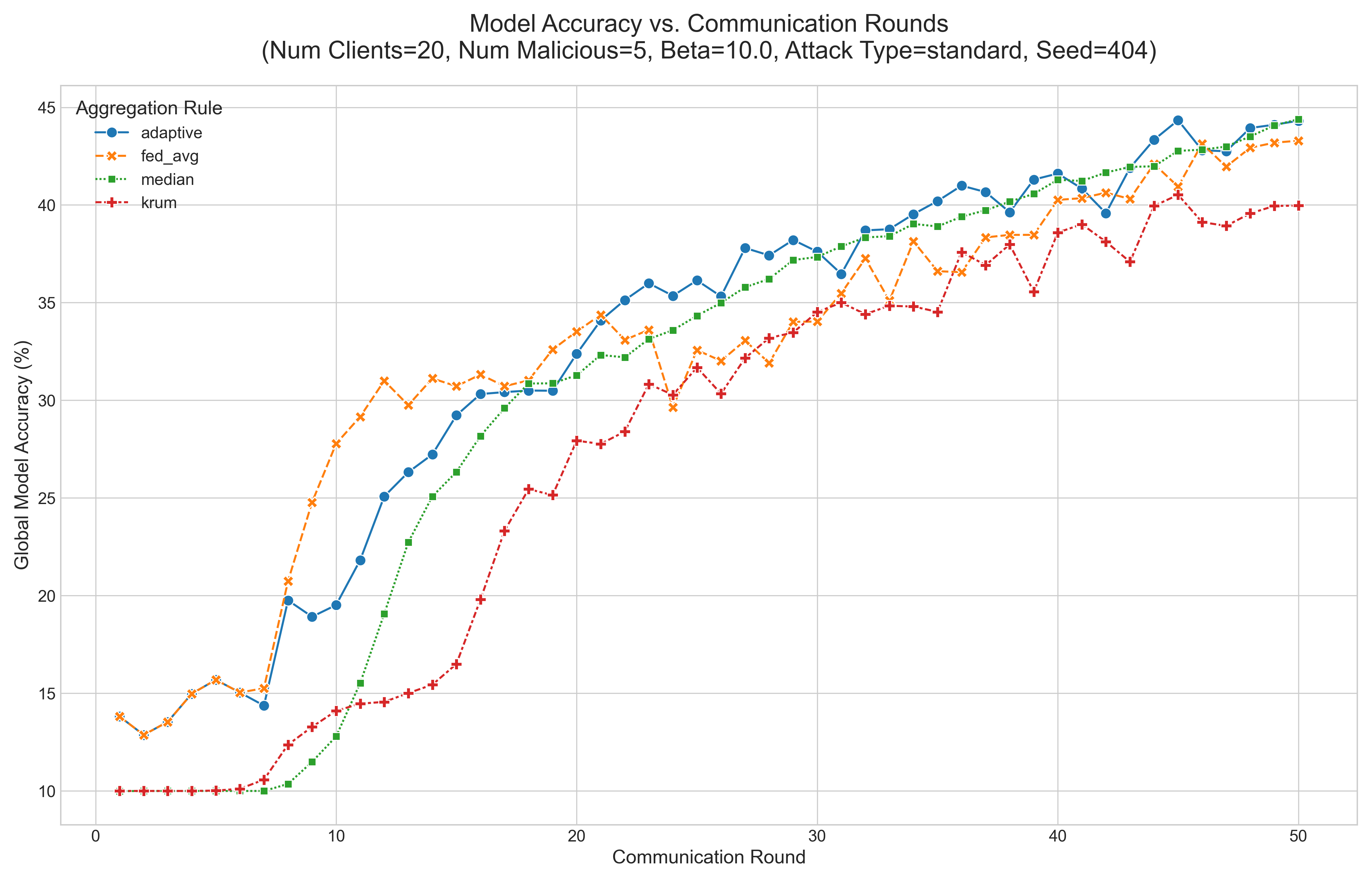}
    \caption{Model accuracy in a Krum-favorable environment ($\beta=10.0$, standard attack). The adaptive agent correctly identifies and tracks the performance of 'Krum' as the optimal static strategy.}
    \label{fig:exp_krum_favorable}
\end{figure}

As shown in Fig.~\ref{fig:exp_krum_favorable}, the results validate our hypothesis perfectly. In this environment, \texttt{Krum} is the clear champion among static rules. The 'adaptive' agent's performance demonstrates its intelligence: after an initial exploration phase, its accuracy curve converges to and tracks the optimal \texttt{Krum} baseline. This proves that FedStrategist is a true generalist, capable of learning the optimal defensive posture based on the specific context of the federated environment.

\subsection{Controllability of the Agent's Risk Posture}
Our final experiment validates the core claim that the $\lambda_{cost}$ parameter in the agent's reward function, $R_t = (\Delta \text{Acc}) - (\lambda \times C_j)$, acts as an interpretable "risk dial." We ran the adaptive simulation under the stealth attack scenario with four different values of $\lambda$.

\begin{figure*}[htbp]
    \centering
    \includegraphics[width=\textwidth]{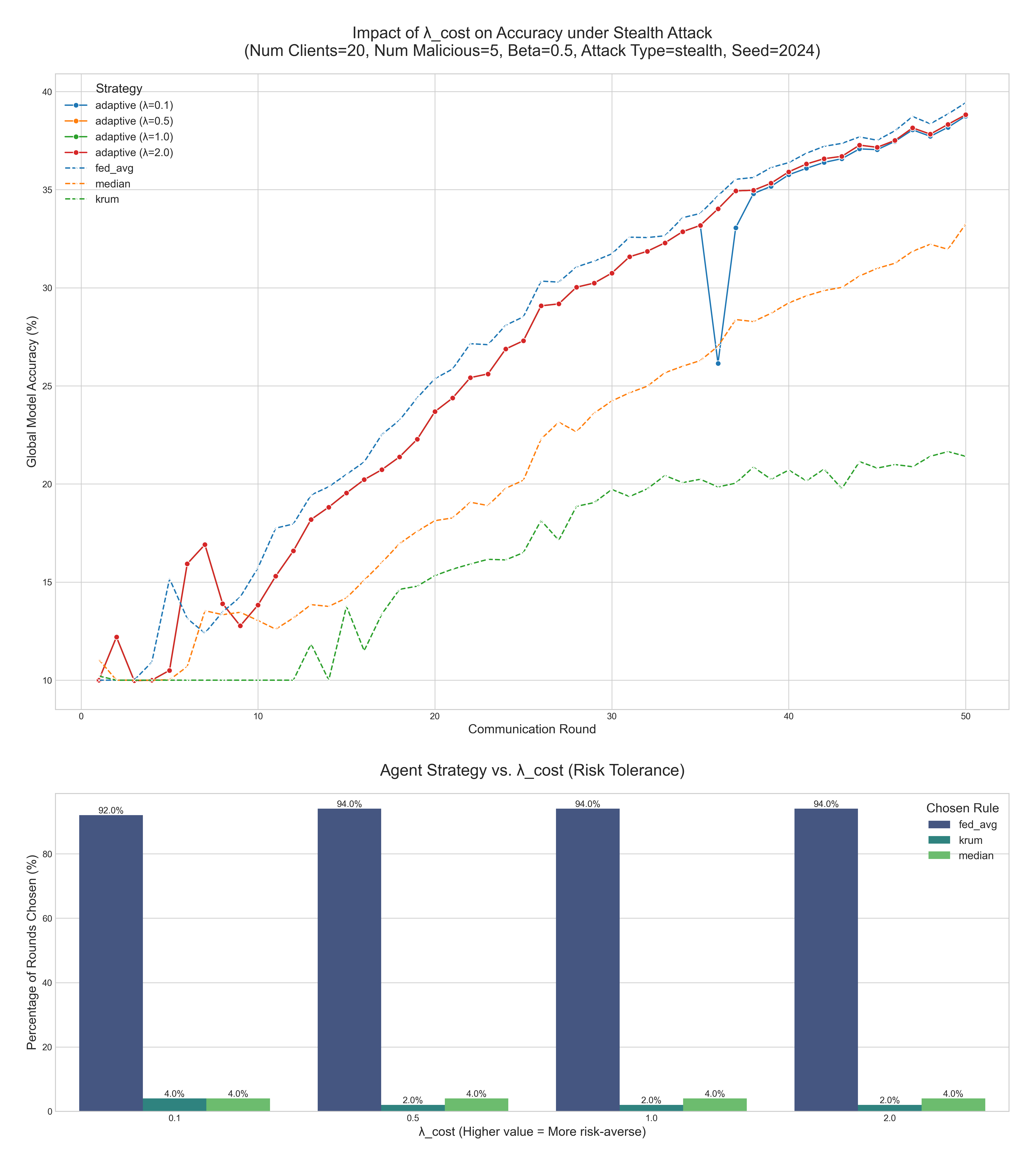}
    \caption{Impact of the $\lambda_{cost}$ parameter on agent performance and strategy under the stealth attack. (Left) Accuracy curves for different $\lambda$ values, demonstrating a clear shift from risk-seeking (tracking 'FedAvg') to more conservative behavior. (Right) The agent's chosen rule percentages, confirming the learned policy change.}
    \label{fig:exp_lambda}
\end{figure*}

The findings, presented in Fig.~\ref{fig:exp_lambda}, are definitive. The $\lambda_{cost}$ parameter directly and predictably controls the agent's behavior.
\begin{itemize}
    \item \textbf{Low $\lambda$ (Risk-Seeking):} When the cost of defense is negligible, the agent learns an aggressive policy. The bar chart shows it selects the high-speed 'FedAvg' over 90\% of the time, and its accuracy curve mirrors the 'FedAvg' baseline.
    \item \textbf{High $\lambda$ (Risk-Averse):} When the cost of defense is high, the agent learns a more conservative policy. Its accuracy curve detaches from the 'FedAvg' baseline, providing a more stable, though slower, learning trajectory, and its selection of robust rules like 'Median' and 'Krum' increases.
\end{itemize}

This experiment confirms that FedStrategist is not a monolithic black box but a \textbf{tunable controller}. A practitioner can use the $\lambda$ parameter to explicitly define the system's risk posture, providing a practical and powerful tool for deploying federated learning in real-world environments with varying security constraints.

\section{Discussion}
\label{sec:discussion}

The experimental results presented in Section~\ref{sec:results} not only confirm the efficacy of our adaptive framework, FedStrategist, but also illuminate the complex interplay between data distribution, adversarial strategy, and defensive posture in Federated Learning. This section interprets these findings, discusses their implications for designing robust AI systems, and solidifies the contributions of our work.

\subsection{The Brittleness of Static Defenses and the Need for Adaptation}
Our findings provide strong empirical evidence that no single static aggregation rule is a panacea for securing Federated Learning. The performance of a given rule is highly contingent on the specific context of the learning environment. As demonstrated, \texttt{Krum}, a geometrically-motivated defense, excels when its core assumptions hold—namely, when benign clients form a tight geometric cluster and adversaries are clear outliers (Fig.~\ref{fig:exp_krum_favorable}). However, these assumptions are fragile; in the presence of high data heterogeneity ($\beta=0.1$), the benign clients are naturally dispersed, and \texttt{Krum}'s performance collapses entirely (Fig.~\ref{fig:exp_standard_attack}).

This context-dependency renders any static choice of defense inherently brittle. A system optimized for one threat model may be completely vulnerable to another. Our results show that an adaptive framework like FedStrategist provides a more robust and higher-performing solution precisely because it is not committed to a single doctrine. By learning to select the appropriate tool for the current conditions, it transcends the limitations of its constituent parts.

\subsection{Interpreting the "Paradox of the Losing Winner"}
The outcome of the Stealth Attack experiment (Fig.~\ref{fig:exp_stealth_attack}) presents a fascinating paradox: the completely undefended \texttt{FedAvg} achieved the highest final accuracy. This result underscores a critical consideration for the field: naively optimizing for test-set accuracy is a dangerous path. The \texttt{FedAvg} model did not "win"; it was successfully and subtly \textbf{manipulated}. The stealth attacker, by normalizing its update magnitude, inadvertently regularized its own attack, reducing the per-round damage. \texttt{FedAvg}, being non-robust, incorporated this malicious drift every round, achieving a high score on the test set while learning a fundamentally compromised and untrustworthy model.

Our \texttt{adaptive} agent, in contrast, behaved as an effective risk manager. Its diagnostic metrics correctly identified a persistent attack, and it consequently learned a conservative policy that prioritized model integrity over raw, compromised performance. The slightly lower accuracy of the \texttt{adaptive} agent is not a failure, but rather the measurable "cost of security" paid to produce a reliable final model. This highlights that the objective in secure FL is not merely accuracy, but trustworthy accuracy.

\subsection{The Controllable Risk Posture of FedStrategist}
The agent's conservative behavior is not a fixed limitation but a controllable feature, which represents a core contribution of our framework. The results of the $\lambda_{cost}$ experiment (Fig.~\ref{fig:exp_lambda}) directly address the agent's perceived preference for \texttt{FedAvg} in the stealth scenario. The agent's policy is a direct and rational response to its reward function, which explicitly balances accuracy gains against the cost of defense. When $\lambda$ is low (risk-seeking), the agent's policy converges to that of the high-risk, high-reward \texttt{FedAvg}. When $\lambda$ is high (risk-averse), its policy converges to that of the safer, more stable \texttt{Median}.

This demonstrates that FedStrategist is not a black box but a tunable controller. It reframes the defense problem from a search for a single "best" algorithm to providing a system where a practitioner can define their desired security posture via a single, interpretable parameter. This tunability is a critical feature for deploying FL systems in real-world applications with varying risk requirements.

\subsection{The Sufficiency of Lightweight Diagnostics}
Finally, the success of the agent in the Stealth Attack scenario is a powerful demonstration that a multi-faceted, lightweight state vector can provide sufficient information for robust decision-making. Even with one of its primary metrics (\texttt{norm\_variance}) completely neutralized by the adversary, the agent learned an effective policy by leveraging the remaining signals, such as \texttt{cosine\_similarity}. This provides strong evidence that a small vector of diverse, computationally cheap metrics can be more resilient than a single, more complex diagnostic tool, avoiding the need for computationally expensive feature engineering in every round and making the framework practical for real-world deployment.

\section{Conclusion}
\label{sec:conclusion}

In this paper, we addressed the critical challenge of selecting an appropriate defense mechanism in Byzantine-robust Federated Learning. We first established through a comprehensive analysis that no single static aggregation rule is universally optimal; its performance is highly contingent on the specific data distribution and adversarial strategy it faces.

To solve this problem, we designed, implemented, and evaluated \textbf{FedStrategist}, a novel meta-learning framework that treats aggregation rule selection as a real-time control problem. Our empirical results demonstrate the superiority and intelligence of this adaptive approach. We have shown that our contextual bandit agent successfully learns policies that outperform any single static rule across diverse and challenging scenarios. We validated its generality by creating distinct environments where different defenses (\texttt{Median}, \texttt{Krum}) were optimal, and in each case, the agent correctly identified and adopted the superior strategy.

Most significantly, our analysis of the stealth attack scenario revealed a crucial insight: optimizing for raw test accuracy alone is insufficient and can be misleading. We showed that FedStrategist successfully navigates the "paradox of the losing winner," learning a conservative policy to guarantee model integrity even when a compromised, non-robust baseline appeared to perform better. Furthermore, we demonstrated that our framework is not a black box, but a controllable one. By tuning a single, interpretable "risk tolerance" hyperparameter ($\lambda_{cost}$), a practitioner can explicitly guide the agent's policy on a spectrum from aggressive and performance-focused to conservative and security-focused. The primary contribution of this work is therefore a practical, tunable, and analyzable framework for creating more intelligent and truly resilient security for decentralized AI systems.

\section{Future Work}
\label{sec:future_work}

Our research opens several promising avenues for future investigation. While our empirical analysis demonstrates the stability of the FedStrategist agent under the tested conditions, a formal theoretical analysis of the convergence guarantees for a learning system governed by a non-stationary, adaptive aggregation policy remains a significant open problem.

Immediate next steps could focus on expanding the capabilities of the agent. The "Defense Arsenal" could be enriched with a wider array of aggregation rules, including more sophisticated defenses against collusion (e.g., \texttt{FoolsGold}) or attacks specifically targeting non-IID data. Similarly, the diagnostic "State Vector" could be enhanced with more nuanced, information-rich metrics. A particularly interesting direction would be to investigate the agent's resilience to adversaries that directly attempt to manipulate multiple state metrics simultaneously, which would test the limits of our lightweight diagnostic approach.

Finally, this work represents a step towards a new paradigm of self-aware, self-defending decentralized AI systems. The one-way communication in our framework (from diagnostics to agent) could be made bi-directional. A long-term research goal is the full co-design of learning and security protocols, where the FL algorithm itself could adapt its training parameters based on signals from the defense agent, creating a truly symbiotic and resilient ecosystem.

\section*{Supporting Information}

This document provides supplementary materials to support the findings presented in the main text. This includes detailed numerical results from all experimental runs, round-by-round strategy selections made by our adaptive agent, and a description of the framework's hyperparameters.

\paragraph*{S1 Table. Aggregated Performance Metrics for All Experimental Scenarios.}
\label{S1_Table}
{\bf Final accuracy and standard deviation of accuracy across the final 10 rounds.} The following tables provide the final performance metrics for each aggregation strategy under the three distinct experimental scenarios. The "Final Accuracy" is the accuracy recorded at the final communication round (Round 50 or 75). "Std Dev (Last 10)" measures the standard deviation of the accuracy over the last 10 rounds, serving as a proxy for the stability of the learned model.

\begin{table}[htbp]
\centering
\caption{Scenario 1: High Heterogeneity ($\beta=0.1$), Standard Attack}
\begin{tabular}{lcc}
\toprule
\textbf{Aggregation Rule} & \textbf{Final Accuracy (\%)} & \textbf{Std Dev (Last 10)} \\
\midrule
FedAvg & 20.17 & 5.86 \\
Median & 27.38 & 0.89 \\
Krum & 10.00 & 0.45 \\
Adaptive ($\lambda=0.5$) & \textbf{28.73} & \textbf{0.34} \\
\bottomrule
\end{tabular}
\end{table}

\begin{table}[htbp]
\centering
\caption{Scenario 2: Moderate Heterogeneity ($\beta=0.5$), Stealth Attack}
\begin{tabular}{lcc}
\toprule
\textbf{Aggregation Rule} & \textbf{Final Accuracy (\%)} & \textbf{Std Dev (Last 10)} \\
\midrule
FedAvg & \textbf{39.43} & 1.25 \\
Median & 33.25 & 0.81 \\
Krum & 21.41 & 0.72 \\
Adaptive ($\lambda=0.5$) & 38.82 & \textbf{0.48} \\
\bottomrule
\end{tabular}
\end{table}

\begin{table}[htbp]
\centering
\caption{Scenario 3: Low Heterogeneity ($\beta=10.0$), Standard Attack}
\begin{tabular}{lcc}
\toprule
\textbf{Aggregation Rule} & \textbf{Final Accuracy (\%)} & \textbf{Std Dev (Last 10)} \\
\midrule
FedAvg & 43.29 & 1.04 \\
Median & \textbf{44.39} & \textbf{0.57} \\
Krum & 39.97 & 0.65 \\
Adaptive ($\lambda=0.5$) & 44.31 & 0.88 \\
\bottomrule
\end{tabular}
\end{table}

\paragraph*{S2 Table. Agent Strategy Selection vs. Risk Tolerance ($\lambda$).}
\label{S2_Table}
{\bf Percentage of rounds each aggregation rule was chosen.} This table details the learned policy of the FedStrategist agent under the "Stealth Attack" scenario ($\beta=0.5$) for different values of the risk-tolerance parameter, $\lambda_{cost}$. The data shows a clear shift in policy from risk-seeking (favoring the low-cost 'FedAvg') to risk-averse (exploring safer, more expensive rules) as $\lambda$ increases.

\begin{table}[htbp]
\centering
\caption{Agent Policy Distribution vs. $\lambda_{cost}$ (Stealth Attack, 75 Rounds)}
\begin{tabular}{lccc}
\toprule
\textbf{$\lambda_{cost}$} & \textbf{\% FedAvg} & \textbf{\% Median} & \textbf{\% Krum} \\
\midrule
0.1 (Risk-Seeking) & 92.0\% & 4.0\% & 4.0\% \\
0.5 (Balanced) & 94.0\% & 4.0\% & 2.0\% \\
1.0 (Risk-Averse) & 94.0\% & 4.0\% & 2.0\% \\
2.0 (Highly Risk-Averse) & 94.0\% & 4.0\% & 2.0\% \\
\bottomrule
\end{tabular}
\end{table}

\paragraph*{S1 Appendix. Hyperparameter and Framework Details.}
\label{S1_Appendix}
{\bf Full experimental parameters for reproducibility.} To ensure the reproducibility of our results, we provide the complete set of hyperparameters used in our simulation framework.
\begin{itemize}
    \item \textbf{Dataset:} CIFAR-10, partitioned using a Dirichlet distribution with $\beta \in \{0.1, 0.5, 10.0\}$.
    \item \textbf{Model Architecture:} A simple CNN with two convolutional layers (3x3 kernel, ReLU activation, Max-pooling) and three fully-connected layers.
    \item \textbf{FL Training:} $N=20$ clients, of which $f=5$ are malicious. Local training is performed for $E=1$ epoch using SGD with a learning rate of 0.001 and momentum of 0.9. Batch size is 32.
    \item \textbf{Bandit Agent (LinUCB):} The exploration parameter was set to $\alpha=1.5$. The state vector was composed of (1) variance of L2 norms, (2) avg. pairwise cosine similarity, and (3) norm of the mean update vector.
    \item \textbf{Reward Function:} $R_t = (\text{Acc}_{t} - \text{Acc}_{t-1}) - (\lambda \times C_j)$. Heuristic costs were assigned as $C_{FedAvg}=0.1$, $C_{Median}=0.4$, and $C_{Krum}=0.8$. The risk-tolerance parameter $\lambda_{cost}$ was varied in $\{0.1, 0.5, 1.0, 2.0\}$.
\end{itemize}

\paragraph*{S1 File. Simulation Framework Source Code.}
\label{S1_File}
{\bf Complete Python source code.} A zipped archive containing the complete Python source code for our custom simulation framework, 'FedStrategist', as well as the shell scripts used to run the experiments and the Python script to generate the plots, ensuring full reproducibility of our findings. The code is available \href{https://github.com/rafidhaque/FedStrategist}{here}.


\nolinenumbers

\section*{Data Availability Statement}
\label{sec:data_availability}

All relevant data and source code are available to ensure the reproducibility of our findings. The raw CSV log files for all experimental runs presented in this study, the Python scripts used to generate the plots, and the complete source code for the 'FedStrategist' simulation framework are permanently archived and publicly accessible on Zenodo. The dataset can be accessed at \url{https://doi.org/10.5281/zenodo.16068113}, under DOI: 10.5281/zenodo.16068113. The development repository is also maintained on GitHub at \url{https://github.com/rafidhaque/FedStrategist}.

\bibliography{references}

\end{document}